\documentclass[journal,twoside,web]{ieeecolor}
\usepackage{tmi}
\usepackage{cite}
\usepackage{amsmath,amssymb,amsfonts}
\usepackage{algorithmic}
\usepackage{graphicx}
\usepackage{textcomp}

\usepackage{makecell}
\usepackage{url}
\usepackage{booktabs}
\usepackage{multirow}
\newcommand*{\pcr}{\fontfamily{pcr}\selectfont}
\usepackage{mathtools}
\usepackage{colortbl}
\definecolor{colorlightblue}{rgb}{0.878, 1, 1}
\definecolor{green}{rgb}{0, 0.698, 0}
\definecolor{gray}{rgb}{0.6, 0.6, 0.6}

\def\BibTeX{{\rm B\kern-.05em{\sc i\kern-.025em b}\kern-.08em
    T\kern-.1667em\lower.7ex\hbox{E}\kern-.125emX}}
\begin{document}
\title{Efficient Medical Vision-Language Alignment Through Adapting Masked Vision Models}
\author{Chenyu Lian, Hong-Yu Zhou, Dongyun Liang, Jing Qin, \IEEEmembership{Senior Member, IEEE}, and Liansheng Wang, \IEEEmembership{Member, IEEE}
\thanks{This work was supported by National Natural Science Foundation of China (Grant No. 62371409), Fujian Provincial Natural Science Foundation of China (Grant No. 2023J01005), Innovation and Technology Fund under Hong Kong Innovation and Technology Commission (Project No. ITS/202/23), Collaborative Research with World-leading Research Groups in The Hong Kong Polytechnic University (Project No. G-SACF), and a Shenzhen-Hong Kong-Macao Science and Technology Plan Project (Category C Project) under Shenzhen Municipal Science and Technology Innovation Commission (project no SGDX20230821092359002).}
\thanks{Chenyu Lian is with the School of Informatics, Xiamen University, Xiamen 361005, China, and the Center for Smart Health, School of Nursing, the Hong Kong Polytechnic University, Hong Kong, China (e-mail:chenyu.lian@connect.polyu.hk).}
\thanks{Hong-Yu Zhou is with the Department of Biomedical Informatics, Harvard
Medical School, Boston, USA (e-mail: whuzhouhongyu@gmail.com).}
\thanks{Dongyun Liang is with the Department of Radiology, Zhongshan Hospital (Xiamen), Fudan University, Xiamen Municipal Clinical Research Center for Medical Imaging, Fujian Province Key Clinical Specialty for Medical Imaging, Xiamen Key Laboratory of Clinical Transformation of Imaging Big Data and Artificial Intelligence, Xiamen 361015, China (email: ldy7372@163.com).}
\thanks{Liansheng Wang is with the National Institute for Data Science
in Health and Medicine, and the Department of Computer Science,
School of Informatics, Xiamen University, Xiamen 361005, China (e-mail:
lswang@xmu.edu.cn).}
\thanks{Jing Qin is with the Center for Smart Health, School of Nursing,
The Hong Kong Polytechnic University, Hong Kong, China (e-mail:
harry.qin@polyu.edu.hk).}
\thanks{Dongyun Liang and Liansheng Wang are the corresponding authors.}
}
\maketitle
\begin{abstract}
Medical vision-language alignment through cross-modal contrastive learning shows promising performance in image-text matching tasks, such as retrieval and zero-shot classification.
However, conventional cross-modal contrastive learning (CLIP-based) methods suffer from suboptimal visual representation capabilities, which also limits their effectiveness in vision-language alignment.
In contrast, although the models pretrained via multimodal masked modeling struggle with direct cross-modal matching, they excel in visual representation.
To address this contradiction, we propose ALTA (ALign Through Adapting), an efficient medical vision-language alignment method that utilizes only about 8\% of the trainable parameters and less than 1/5 of the computational consumption required for masked record modeling.
ALTA achieves superior performance in vision-language matching tasks like retrieval and zero-shot classification by adapting the pretrained vision model from masked record modeling.
Additionally, we integrate temporal-multiview radiograph inputs to enhance the information consistency between radiographs and their corresponding descriptions in reports, further improving the vision-language alignment.
Experimental evaluations show that ALTA outperforms the best-performing counterpart by over 4\% absolute points in text-to-image accuracy and approximately 6\% absolute points in image-to-text retrieval accuracy.
The adaptation of vision-language models during efficient alignment also promotes better vision and language understanding.
Code is publicly available at \url{https://github.com/DopamineLcy/ALTA}.
\end{abstract}


\section{Introduction}
\label{sec:intro}
\begin{figure*}[!t]
  \centering
  \includegraphics[width=\linewidth]{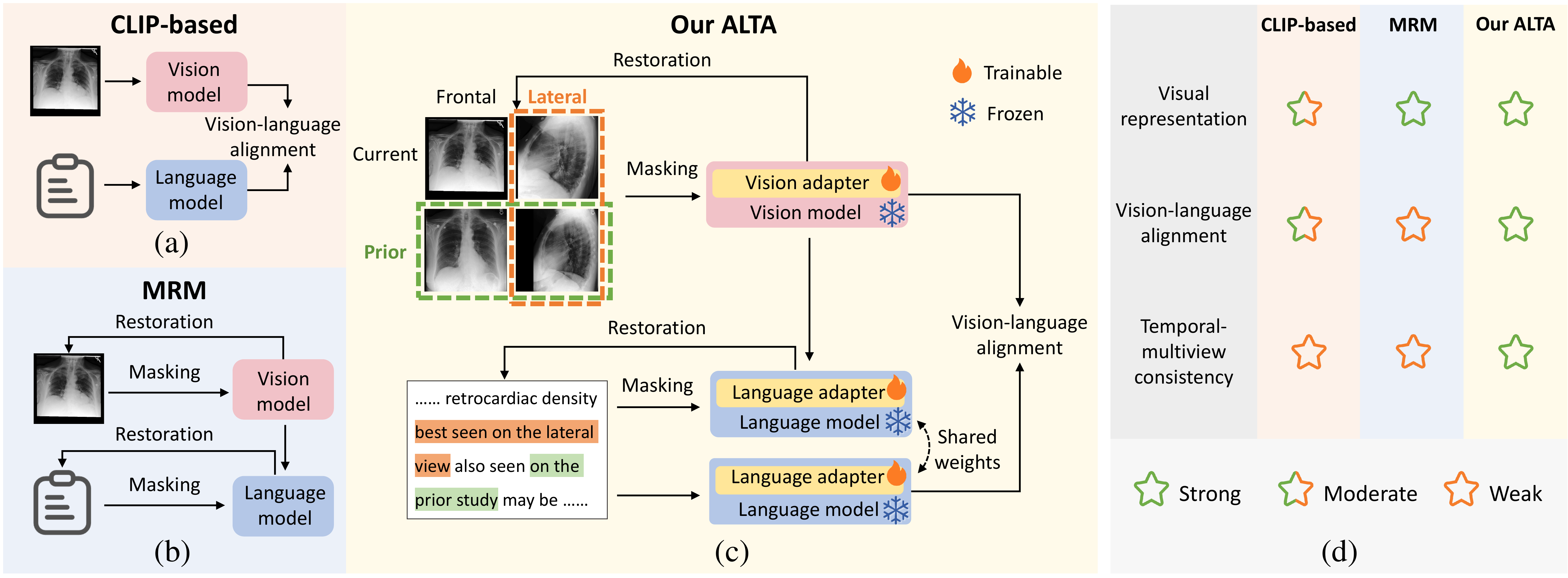}
  \caption{Comparison of CLIP-based methods, MRM, and our ALTA.
  We integrate temporal-multiview inputs and execute efficient vision-language alignment by introducing trainable adapters.
  Radiographs from the prior study and lateral view are integrated to match radiology reports better.
  }
  \label{fig: introduction}
\end{figure*}
\IEEEPARstart{V}{ision}-language models significantly advance medical image representations and emerge as the preferred alternative to pretrained models on ImageNet~\cite{deng2009imagenet} or large radiograph datasets~\cite{johnson2019mimic,wang2017chestx,irvin2019chexpert}.
Effective vision encoder is commonly obtained by CLIP-based~\cite{radford2021learning} image-text contrastive learning~\cite{zhang2022contrastive,huang2021gloria,boecking2022making,zhang2025multimodal}~(see Fig.~\ref{fig: introduction}~(a)) or masked record modeling (MRM)~\cite{zhou2023advancing}~(see Fig.~\ref{fig: introduction}~(b)), which improves the performance in downstream tasks like classification and segmentation.
Benefiting from vision-language contrastive learning, the CLIP-based models are naturally suitable to solve vision-language matching tasks like retrieval and zero-shot classification.
However, it falls short in visual representation, consequently limiting medical vision-language alignment.
In contrast, the pretrained masked vision model of MRM produces better visual representations, which exhibits notable advantages over CLIP-based methods~\cite{zhou2023advancing, perez2024rad, shrestha2023medical, dai2024unichest, liu2023improving, perez2025exploring}.
Nevertheless, the lack of cross-modal alignment impedes MRM from tackling vision-language matching tasks.
This inherent contradiction raises a natural question: \textbf{\textit{How can we adapt the masked vision models to achieve vision-language alignment?}}

On the other hand, radiology reports frequently include descriptions of both longitudinal changes and information from different views.
Since a single patient typically undergoes multiple radiology studies, many radiology examinations include radiographs from multiple perspectives, commonly including frontal and lateral views.
Neglecting this temporal and multiview information leads to inconsistencies between radiographs and their corresponding reports, hindering medical vision-language alignment.
Although some previous studies adopt either temporal or multiview radiograph inputs~\cite{bannur2023learning, zhou2022generalized, karwande2022chexrelnet, kim2023chexfusion, liu2021act, wu2022deltanet}, the impact on vision-language alignment is largely overlooked, and these two factors are not considered simultaneously.
The observation prompts another valuable question: \textbf{\textit{How can we integrate temporal-multiview radiograph inputs to enhance vision-language alignment?}}

To address the two questions posed, we introduce ALTA (\textbf{AL}ign \textbf{T}hrough \textbf{A}dapting), a parameter-efficient method that adapts the pretrained vision model from masked record modeling for medical vision-language alignment with only 8\% trainable parameters and 1/5 computational consumption compared to MRM.
As illustrated in Fig.~\ref{fig: introduction}~(c), the visual representations from MRM are further aligned with language representations efficiently through optimizing the integrated adapters.
In addition, the temporal-multiview radiograph inputs enhance vision-language alignment by improving the information consistency between paired radiographs and radiology reports.

In practice, we adopt the masked vision model of MRM and a model for biomedical language understanding~\cite{boecking2022making} as the vision and language models, respectively.
Trainable adapters are integrated into the frozen vision and language models, comprising only 8\% of the total parameters.
After restructuring radiology records, we align the representations between temporal-multiview radiographs and radiology reports using contrastive losses.
As a result, our ALTA outperforms baseline methods in three retrieval tasks by a large margin.
For example, improvements of over 4\% absolute points and approximately 6\% absolute points can be observed for the accuracy scores of text-to-image retrieval and image-to-text retrieval, respectively.
ALTA consistently surpasses all counterparts by notable margins on radiograph zero-shot classification and sentence similarity analysis tasks.
The results of downstream classification tasks verify the enhanced visual representation capability after the adaptation.
Finally, comprehensive ablation analyses validate the efficacy and necessity of the proposed temporal-multiview inputs and adopted loss functions.

This work pioneers in adopting vision-language alignment as a separate and efficient stage through adaptation rather than necessary supervision during pretraining.
We hope to reduce the burden on pretraining and inspire more research on efficient alignment based on advanced representations.
\section{Related Work}
\subsection{Masked Record Modeling for Radiograph Representation Learning}
Masked record modeling (MRM)~\cite{zhou2023advancing} verifies its advantage over both CLIP-based report-supervised methods~\cite{zhang2022contrastive, boecking2022making, zhou2022generalized} and self-supervised methods~\cite{chen2019self, zhou2021models, zhou2020comparing} in visual representation.
Specifically, MRM restores masked radiograph patches using non-masked ones and restores masked report tokens from vision-language hybrid representations, producing superior performance in downstream tasks.
However, the lack of vision-language alignment limits its direct applicability for retrieval and zero-shot classification tasks.
To overcome the limitations of MRM, the introduced parameter-efficient method adapts the effective pretrained vision encoder to achieve medical vision-language alignment.
\subsection{Medical Vision-Language Alignment Through Cross-Modal Contrastive Learning}
Vision-language contrastive learning (i.e., CLIP-based) methods explicitly align cross-modal representations, enabling direct vision-language matching without further adaptation~\cite{radford2021learning, zhang2022contrastive, huang2021gloria, zhou2022generalized, li2022blip, kim2021vilt, li2021align, wang2022image, zhang2023knowledge, lai2024carzero, huang2024enhancing}.
However, the visual representation capability of the pretrained encoder from cross-modal contrastive learning lags behind the masked modeling approaches due to insufficient vision semantic modeling, consequently limiting vision-language alignment.
Despite advancements such as local alignment, autoregressive report modeling, masked language modeling, and knowledge enhancement~\cite{huang2021gloria,boecking2022making,zhou2022generalized,wu2023medklip}, MRM remains a simpler, steadier, and more effective pretraining method for radiograph representation learning, exhibiting notable advantages over CLIP-based methods~\cite{perez2024rad}.
Unlike CLIP-based methods, we do not take vision-language contrastive learning as the primary proxy task for visual representation acquisition but as an additional vision-language alignment stage parameter-efficiently.
We leverage the pretrained vision model from MRM and align vision-language representations by conducting cross-modal contrastive learning while incorporating temporal-multiview inputs for better vision-language information consistency.
\begin{figure*}[!t]
  \centering
  \includegraphics[width=\textwidth]{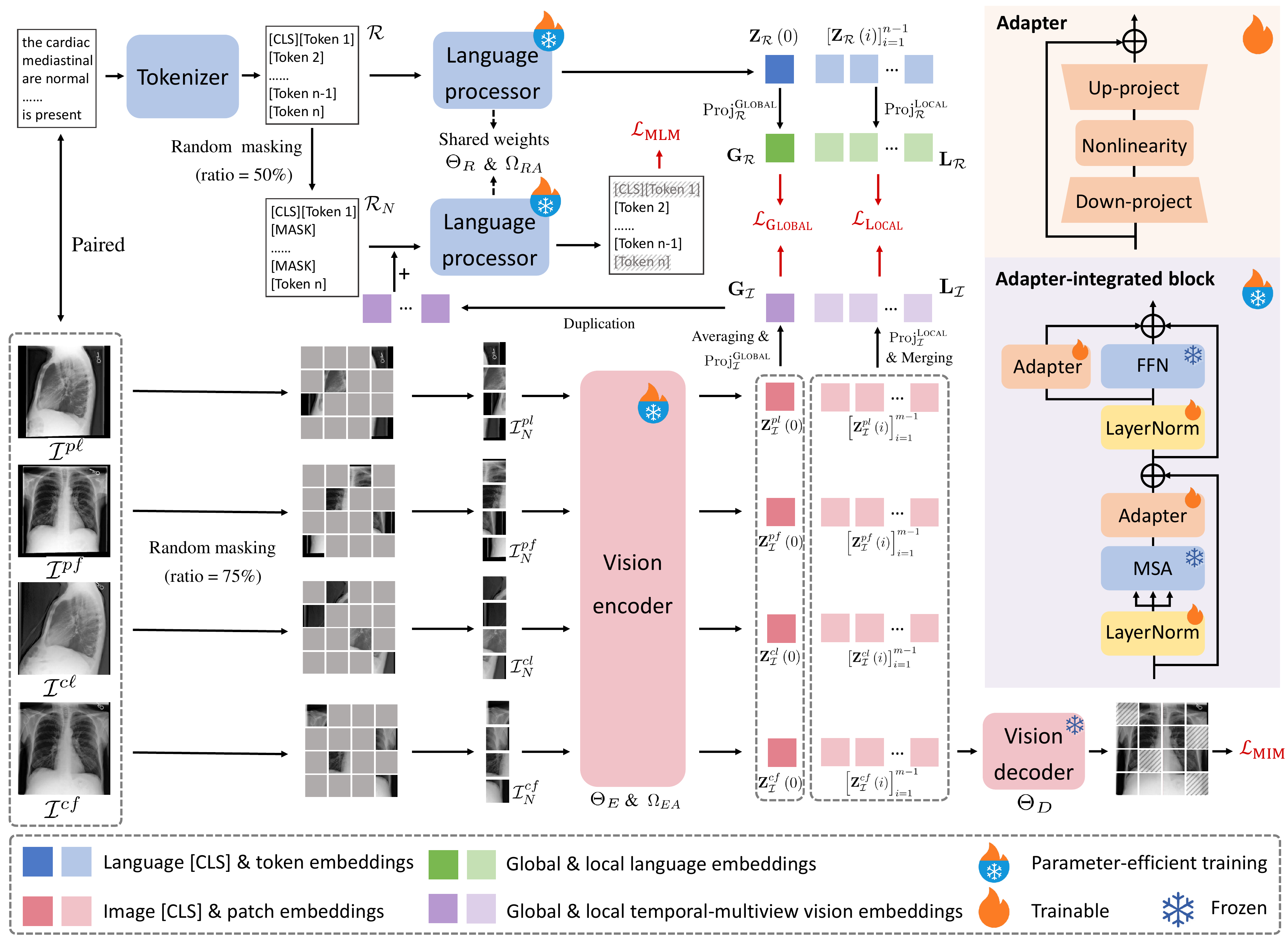}
  \caption{Overview of ALTA and adapter integration.
  The illustrated adapter-integrated block is employed as the transformer block in both the vision encoder and language processor.
  Only 8\% of the trainable parameters are included in the framework during efficient vision-language alignment.
  Global and local alignment losses are adopted to align vision-language representations while MLM and MIM losses contribute to the maintenance of masked modeling capability.
  }
  \label{fig: overview}
\end{figure*}
\subsection{Parameter-Efficient Training}
Parameter-efficient training methods that are originally proposed for the fine-tuning stage in NLP to reduce computation and storage costs~\cite{houlsby2019parameter, hu2021lora, liu2022few, zaken2022bitfit, he2021towards} have been adopted in computer vision~\cite{jia2022visual, sung2022vl, yang2023aim, chen2022vision} and, more recently, in medical image analysis to enhance downstream task performance~\cite{dutt2023parameter, zhu2023melo, lian2024less, fischer2024prompt, silva2023towards}.
Khan \textit{et al.}~\cite{khan2023contrastive} employ parameter-efficient training for contrastive learning between visual and language data based on ImageNet pretrained weights.
Our ALTA adapts the advanced radiograph vision encoder from masked record modeling to enable parameter-efficient medical vision-language alignment.
\section{Efficient Vision-Language Alignment Through Adaptation}
\label{sec: method}
We propose ALTA, an efficient alignment method that aligns the representations between radiographs and radiology reports.
The motivation behind this is twofold.
First, while masked record modeling~\cite{zhou2023advancing} offers a strong vision encoder, its weak vision-language alignment restricts its application in vision-language matching.
Second, radiographs from the prior study and lateral view often contain supplementary information frequently referenced in radiology reports, which are often overlooked for vision-language alignment.
In contrast, the proposed ALTA not only aligns vision-language representations by adapting the masked modeling pretrained model efficiently but also highlights the consistency of temporal-multiview information between radiographs and radiology reports.

%
The overview of ALTA is illustrated by Fig.~\ref{fig: overview}.
Each input record consists of a radiograph quaternion $(\mathcal{I}^{cf}, \mathcal{I}^{c\ell}, \mathcal{I}^{pf}, \mathcal{I}^{p \ell})$, denoting the \textit{current frontal, current lateral, prior frontal, }and \textit{prior lateral} radiographs, respectively, and a tokenized radiology report $\mathcal{R}$.
It should be noted that $\mathcal{I}^{c\ell}, \mathcal{I}^{pf}$, and $\mathcal{I}^{p\ell}$ may be missing, in which case they are filled with zeros.
In the ALTA framework, we first apply random masking to each radiograph by removing 75\% of its patches for efficiency and to enable masked modeling tasks~\cite{zhou2023advancing, li2023scaling}.
Then the non-masked patches are forwarded to the vision encoder to generate visual representations, which are then aggregated into global and local temporal-multiview vision embeddings.
Different from radiographs, the tokenized radiology report is forwarded to the language processor directly without being masked, generating global and local language embeddings.
Subsequently, the vision-language embeddings are aligned through global and local alignment losses.

To maintain the capability of representation for masked modeling, ALTA includes MLM (masked language modeling) and MIM (masked image modeling) losses.
In practice, for MLM loss, we randomly mask 50\% tokens of the tokenized report by replacing them with the special token {\pcr{[MASK]}} and obtain hybrid embeddings by adding global temporal-multiview vision embeddings to each token embedding, which are passed to the language processor for masked language restoration.
For MIM loss, visual representations of the current frontal radiograph are fed into the image decoder for super-resolution restoration.
\subsection{Trainable Adapters Integration}
We employ adapter~\cite{houlsby2019parameter}, a simple yet effective parameter-efficient learning module, to serve as the integrated trainable components of ALTA.
Fig.~\ref{fig: overview} depicts the architecture of the adapters and their integration strategy.
Both the vision encoder and language processor are based on Transformer~\cite{vaswani2017attention}, and we incorporate adapters twice within each Transformer block: following the multi-head self-attention (MSA) layer and in parallel to the feed-forward networks (FFN) layer.
An adapter includes a down-project linear layer that reduces the input to a lower dimension, followed by a non-linear GELU layer~\cite{hendrycks2016gaussian}, and an up-project linear layer to restore the latent feature to the original dimension.
\subsection{Temporal-Multiview Radiograph Representations}
Define the set of temporal-view types $\mathcal{S} \coloneqq \left\{cf, c\ell, pf, p\ell\right\}$.
To introduce the position and temporal-view type information, we add fixed (i.e., unlearnable) positional embeddings and learnable temporal-view embeddings to the radiographs.
Suppose each input radiograph consists of a masked set and a non-masked set, i.e., $\mathcal{I}^{s} = \mathcal{I}^{s}_{M} \sqcup \mathcal{I}^{s}_{N}, \forall s \in \mathcal{S}$.
The non-masked sets are then fed into the shared vision encoder to generate the temporal-multiview radiograph representations\footnote{Position and temporal-view embeddings are omitted for simplicity.}:
\begin{equation}
\label{eq:representation_obtain}
\mathbf{Z}_{\mathcal{I}}^{s} = f_{{\rm {\rm \Theta}}_E}\left(\mathcal{I}^{s}_{N};\ {\rm \Omega}_{EA}\right), \forall s \in \mathcal{S}
 \end{equation}
where $\mathbf{Z}_{\mathcal{I}}^{s} = \left[\mathbf{Z}_{\mathcal{I}}^s \left(i\right)\right]_{i=0}^{m-1} \in \mathbb{R}^{m \times d_{\mathcal{I}}}$, $\mathbf{Z}_{\mathcal{I}}^s \left(0\right)$ and $\left[\mathbf{Z}_{\mathcal{I}} ^s\left(i\right)\right]_{i=1}^{m-1}$ stand for the embeddings of the {\pcr{[CLS]}} token and other non-masked image patches, respectively.
$m$ is the number of non-masked patches in each radiograph and $d_{\mathcal{I}}$ denotes the dimension of vision features.
${\rm \Theta}_E$ and ${\rm \Omega}_{EA}$ denote the parameters of the frozen vision model and trainable adapters in the vision encoder, respectively.
\subsection{Efficient Vision-Language Alignment}
We have acquired temporal-multiview radiograph representations through~\eqref{eq:representation_obtain}, which are intended to be aligned with the report representations.
The tokenized radiology report is fed into the language processor to generate latent embeddings:
\begin{equation}
\label{eq:report_representation_obtain}
\mathbf{Z}_{\mathcal{R}} = f_{{\rm \Theta}_R}(\mathcal{R};\ {\rm \Omega}_{RA}),
\end{equation}
where $\mathbf{Z}_{\mathcal{R}} = \left[\mathbf{Z}_{\mathcal{R}}\left(i\right)\right]_{i=0}^{n-1} \in \mathbb{R}^{n \times d_{\mathcal{R}}}$, $\mathbf{Z}_{\mathcal{R}}(0)$ and $\left[\mathbf{Z}_{\mathcal{R}}\left(i\right)\right]_{i=1}^{n-1}$ stand for the embeddings of the {\pcr{[CLS]}} token and other language tokens, respectively.
$n$ is the number of tokens in each report and $d_{\mathcal{R}}$ represents the dimension of language features.
${\rm \Theta}_R$ and ${\rm \Omega}_{RA}$ represent the parameters of the frozen language model and trainable adapters in the language processor, respectively.

Then, report embeddings $\mathbf{Z}_{\mathcal{R}}$ are projected to produce global and local language representations:
\begin{subequations}
\begin{gather}
\mathbf{G}_{\mathcal{R}} = {\rm Proj}_{\mathcal{R}}^{\textsc{Global}}\left(\mathbf{Z}_{\mathcal{R}}\left(0\right)\right),\\
\mathbf{L}_{\mathcal{R}}\left(i\right) = {\rm Proj}_{\mathcal{R}}^{\textsc{Local}}(\mathbf{Z}_{\mathcal{R}}\left(i\right)), \forall i \in \left\{ 1,\cdots,n-1\right\}, \\
\mathbf{L}_{\mathcal{R}} = {\left[ \mathbf{L}_{\mathcal{R}}\left(i\right) \right]}_{i=1}^{n-1},
\end{gather}
\end{subequations}
where $\mathbf{G}_{\mathcal{R}} \in \mathbb{R}^{1 \times d_{\textsc{Global}}}$ and $\mathbf{L}_{\mathcal{R}} \in \mathbb{R}^{\left(n - 1\right) \times d_{\textsc{Local}}}$, $d_{\textsc{Global}}$ and $d_{\textsc{Local}}$ are the dimensions of global and local representations, respectively.
All projectors above and below, i.e., ${\rm Proj}_{\mathcal{R}}^{\textsc{Global}}$, ${\rm Proj}_{\mathcal{R}}^{\textsc{Local}}$, ${\rm Proj}_{\mathcal{I}}^{\textsc{Global}}$, and ${\rm Proj}_{\mathcal{I}}^{\textsc{Local}}$, are MLPs.
Similarly, we can obtain global and local representations of the temporal-multiview radiograph inputs:
\begin{subequations}
\begin{gather}
\mathbf{G}_{\mathcal{I}} = {\rm Proj}_{\mathcal{I}}^{\textsc{Global}}\left(\frac{1}{\left|\mathcal{S}\right|}\sum_{s \in \mathcal{S}} \mathbf{Z}_{\mathcal{I}}^{s}(0)\right),\\
\mathbf{L}^{s}_{\mathcal{I}}\left(i\right) = {\rm Proj}_{\mathcal{I}}^{\textsc{Local}}(\mathbf{Z}^{s}_{\mathcal{I}}(i)), \forall i \in \left\{ 1,\cdots,m-1\right\}, s \in \mathcal{S},\\
\nonumber
\mathbf{L}_{\mathcal{I}} = {{\left[ \mathbf{L}^{cf}_{\mathcal{I}}\left(i\right) \right]}_{i=1}^{m-1}} \oplus {{\Big[ \mathbf{L}^{c\ell}_{\mathcal{I}}\left(i\right) \Big]}_{i=1}^{m-1}} \\
{\oplus} {{\left[ \mathbf{L}^{pf}_{\mathcal{I}}\left(i\right) \right]}_{i=1}^{m-1}} \oplus {{\left[ \mathbf{L}^{p\ell}_{\mathcal{I}}\left(i\right) \right]}_{i=1}^{m-1}},
\end{gather}
\end{subequations}
where  $\mathbf{G}_{\mathcal{I}} \in \mathbb{R}^{1 \times d_{\textsc{Global}}}$ and $\mathbf{L}_{\mathcal{I}}\in \mathbb{R}^{\left({4m - 4}\right) \times d_{\textsc{Local}}}$.
Besides, the operator $\oplus$ denotes the concatenation along the ``N'' (sequence length) dimension of the three dimensions (B, N, C) in Transformers, corresponding to the ``merging'' operation depicted in Fig.~\ref{fig: overview}.

Since the integration of temporal-multiview radiographs does not introduce further difficulty for cross-modal alignment, we choose widely-used contrastive learning that has been validated to be suitable for temporal and multiview data~\cite{bannur2023learning, wang2023mvco} as the training paradigm.
Specifically, we adopt both global and local contrastive loss~\cite{radford2021learning, huang2021gloria} to conduct vision-language alignment based on the widely used InfoNCE loss~\cite{oord2018representation}, presented by two loss functions:
\begin{subequations}
\begin{gather}
\label{eq:global_loss}
\mathcal{L}_{\textsc{Global}} = {\rm InfoNCE}\left(\mathbf{G}_{\mathcal{I}}, \mathbf{G}_{\mathcal{R}}\right),\\
\label{eq:local_loss}
\mathcal{L}_{\textsc{Local}} = {\rm InfoNCE}\left({\rm Attn}\left(\mathbf{L}_{\mathcal{I}}, \mathbf{L}_{\mathcal{R}}\right), \mathbf{L}_{\mathcal{R}}\right),
\end{gather}
\end{subequations}
where ${\rm Attn}(\mathbf{L}_{\mathcal{I}}, \mathbf{L}_{\mathcal{R}}) \in \mathbb{R}^{\left({4m - 4}\right) \times d_{\textsc{Local}}}$ denotes report features-weighted radiograph representations.
\subsection{Masked Record Modeling Maintenance}
We include the masked record modeling losses (i.e., MLM loss and MIM loss) during the alignment, which are considered helpful to the adaptation process for vision-language alignment (see Table~\ref{table:ablation}).
For MLM loss~\eqref{eq:MLM_loss}, assume that each tokenized radiology report consists of a masked set and a non-masked set, i.e., $\mathcal{R}$ = $\mathcal{R}_{M} \sqcup \mathcal{R}_{N}$.
Different from MRM, we adopt temporal-multiview radiograph representations $\mathbf{G}_{\mathcal{I}}$ to generate hybrid representations by duplicating and adding them to each report token embedding:
\begin{equation}
\label{eq:MLM_loss}
\begin{split}
\mathcal{L}_{\textsc{MLM}} =-\sum_{i=1}^{|\mathcal{R}_{M}|} \log {P_{{\rm \Theta}_E,{\rm \Theta}_{R}}}(\mathcal{R}_{M}\left(i\right) \mid\\
\mathcal{R}_{N},\mathbf{G}_{\mathcal{I}};\ {\rm \Omega}_{EA},{\rm \Omega}_{RA}).
\end{split}
\end{equation}
For MIM loss~\eqref{eq:MIM_loss}, we restore the masked patches of the current frontal radiograph from latent representations of non-masked ones, the difference between restored patches and ground truth is measured by mean squared error (MSE):
\begin{equation}
\label{eq:MIM_loss}
\mathcal{L}_{\textsc{MIM}} = \operatorname{MSE}\left(f_{{\rm \Theta}_D}(\mathbf{Z}_{\mathcal{I}}^{cf}),\mathcal{I}^{cf}_{M}\right),
\end{equation}
where ${\rm \Theta}_D$ is the parameters of the frozen image decoder inherited from MRM.

Overall, the efficient adaptation incorporates both vision-language alignment and masked record modeling, where the total multi-task optimization objective function is as follows:
\begin{equation}
\label{eq:Overall_loss}
\mathcal{L} = \mathcal{L}_{\textsc{Global}} + \lambda_1 \mathcal{L}_{\textsc{Local}} + \lambda_2 \mathcal{L}_{\textsc{MLM}} + \lambda_3 \mathcal{L}_{\textsc{MIM}},
\end{equation}
where $\lambda_1$, $\lambda_2$, and $\lambda_3$ are hyperparameters for balancing the loss functions.
\section{Experiments}
We conduct efficient vision-language alignment and evaluate the performance in aligning vision-language representations on comprehensive benchmarks, including three retrieval tasks, two zero-shot classification tasks, and two tasks of radiology language understanding.
Besides, the representation capability is further verified by applying the vision encoder to downstream classification tasks.
\subsection{Temporal-Multiview MIMIC-CXR Dataset Construction for ALTA Training}
The training of ALTA is conducted on MIMIC-CXR~\cite{johnson2019mimic}, one of the largest public Chest X-ray datasets containing radiographs and radiology reports.
Unlike previous work that either discards non-frontal images~\cite{boecking2022making, bannur2023learning} or mixes images of different views together~\cite{zhou2023advancing}, we restructure the records in a temporal-multiview format.
The attribute {\pcr{ViewPosition}} is used to identify the views of radiographs while attributes {\pcr{StudyDate}} and {\pcr{StudyTime}} are used to identify the time for each study.
Specifically, to construct multiview records, we retain a frontal radiograph, a lateral one if available, and the corresponding radiology report for each study.
After that, the frontal and lateral radiographs from the prior study are also included in each record to obtain temporal-multiview records, where each record comprises an image quaternion $\left(\mathcal{I}_{cf}, \mathcal{I}_{c\ell}, \mathcal{I}_{pf}, \mathcal{I}_{p\ell}\right)$ and a radiology report.
After excluding images with undefined views, the restructured MIMIC-CXR dataset contains 21.3k records.
Approximately 75\% of these records contain prior study data, and around 50\% include the corresponding lateral view image.
\subsection{Datasets and Metrics for Evaluation}
\label{subsec: dataset}
To evaluate the performance of vision-language alignment, we conduct comprehensive retrieval tasks, including image-to-image retrieval and text-to-image retrieval evaluated on CheXpert~8$\times$200~\cite{zhang2022contrastive}, and image-to-text retrieval on CheXpert~5$\times$200 dataset~\cite{huang2021gloria}.
The experiments for zero-shot binary and multi-class classification are conducted on CheXpert~5$\times$200 and RSNA Pneumonia~\cite{shih2019augmenting}, respectively.
NIH ChestX-ray~\cite{wang2017chestx} and MS-CXR-T~\cite{bannur2023learning} are utilized to evaluate the vision understanding capability of the pretrained vision encoder by fine-tuning.
To assess the capacity for language understanding, sentence similarity zero-shot classification tasks are conducted on MS-CXR-T~\cite{bannur2023learning} and RadNLI~\cite{miura2020improving}.

\textbf{CheXpert~8$\times$200} includes 8 different abnormality categories commonly found in Chest X-ray images.
Each category comprises 10 query images, 10 sets of radiology texts serving as query texts, and 200 candidate images.
In line with the original paper~\cite{zhang2022contrastive}, we perform image-to-image and text-to-image retrieval tasks on this dataset.
For the former, the model retrieves $k$ images with the highest similarity to the query image, and each retrieved image is verified to be \textit{Right} or \textit{Wrong}, depending on whether its class matches that of the query image or not.
Retrieval precision is adopted for evaluation with $k=5, 10, 50$, denoted as P@5, P@10, and P@50, respectively.
For the latter, we use the same evaluation process and metrics except adopting radiology texts as queries.

\textbf{CheXpert~5$\times$200} comprises 5 common abnormality categories, each containing 200 records.
Each record consists of a radiograph and a set of radiology descriptions written by board-certified radiologists, which can be utilized in image-to-text retrieval and multi-class zero-shot classification.
In the image-to-text retrieval task, we take images as the queries and retrieve texts from the same categories, using P@5, P@10, and P@100 for evaluation.
For multi-class zero-shot classification, assembled prompts from the original paper~\cite{huang2021gloria} are prepared and the task is then transformed to retrieve the prompts from the correct class for each query image, evaluated by accuracy.

\textbf{RSNA Pneumonia} is a dataset for binary classification to distinguish pneumonia or normal radiographs.
The official validation and test splits with 1,500 and 3,000 images are used for the validation during ALTA training and final evaluation, respectively.
Simple prompts ``There is pneumonia'' and ``There is no pneumonia'' are adopted for pneumonia and normal samples, respectively.
We adopt the dataset for the zero-shot classification task of radiographs, where accuracy, F1 and AUC scores are adopted to evaluate the performance.

\textbf{NIH ChestX-ray} defines a multi-label binary classification problem on 14 pathologies, including 112,120 frontal-view chest radiographs split to 70\%/10\%/20\% as training/validation/test sets.
Fine-tuning is conducted in this dataset to evaluate the representation capabilities of pretrained vision models.
We report AUC scores of multi-label binary classification under three labeling ratios: 1\%, 10\%, and 100\%.

\textbf{MS-CXR-T} presents two tasks, including temporal image classification and temporal sentence similarity classification.

For temporal image classification, the dataset includes 1,045 pairs of images classified into three categories: improving, stable, and worsening.
We split it to 70\%/10\%/20\% as training/validation/test sets and conduct fine-tuning in the pretrained vision model to achieve temporal image classification.
The macro accuracy of the three classes is reported to evaluate the results of classification, in line with the original paper.

For temporal sentence similarity classification, the dataset consists of 361 sentence pairs belonging to two categories: paraphrase or contradiction.
Zero-shot binary classification is conducted in the dataset, where accuracy and AUC scores are reported for evaluation.
\begin{table*}[!th]
\caption{
Results for retrieval tasks, in precision (\%).
The \textbf{best} results are in bold, \underline{second-best} results are underlined.
* denotes the results of the methods with unfair advantages in the CheXpert-based benchmarks.
}
\label{table:retrieval}
\centering
\resizebox{1.0\linewidth}{!}{
\begin{tabular}{lcc|cccccc|ccc}
\toprule
\multicolumn{3}{l|}{Dataset} & \multicolumn{6}{c|}{CheXpert 8$\times$200}                                                                                              & \multicolumn{3}{c}{CheXpert 5$\times$200}            \\ \hline
\multicolumn{3}{l|}{Retrieval Task}    & \multicolumn{3}{c|}{Image $\rightarrow$ Image}                                               & \multicolumn{3}{c|}{Text $\rightarrow$ Image}   & \multicolumn{3}{c}{Image $\rightarrow$ Text}    \\ \hline
Method   & Dataset      & Input Size  & P@5           & P@10          & \multicolumn{1}{l|}{P@50}                                  & P@5           & P@10          & P@50          & P@5           & P@10          & P@100         \\ \hline
\rowcolor{colorlightblue} 
Our ALTA  &  MIMIC-CXR  & 224         & \textbf{51.3} & \textbf{47.1} & \multicolumn{1}{l|}{\cellcolor{colorlightblue}\textbf{38.7}} & \textbf{64.5} & \textbf{65.8} & \textbf{53.0} & \textbf{56.8} & \textbf{55.4} & \textbf{46.8} \\ \hline
Random    &  -  & -           & 12.5          & 12.5          & \multicolumn{1}{l|}{12.5}                                  & 12.5          & 12.5.         & 12.5          & 20.0          & 20.0          & 20.0          \\
ImageNet Pretrained  &   ImageNet  & 224         & 14.8          & 14.4          & \multicolumn{1}{l|}{15.0}                                  & -             & -             & -             & -             & -             & -             \\
MRM       &  MIMIC-CXR   & 224         & 26.5          & 25.9          & \multicolumn{1}{l|}{23.3}                                  & -             & -             & -             & -             & -             & -             \\
ConVIRT   &  MIMIC-CXR   & 224         & \underline{45.3}          & \underline{43.0}          & \multicolumn{1}{l|}{\underline{34.3}}                                  & \underline{59.5}          & \underline{57.3}          & \underline{46.5}          & 49.2             & 47.3             & 40.8             \\
BioViL    &   MIMIC-CXR  & 480         & 35.0          & 34.0          & \multicolumn{1}{l|}{29.5}                                  & 38.0          & 41.0          & 39.5          & 38.0          & 41.0          & 39.5          \\
BioViL-T   &  MIMIC-CXR  & 448         & 35.8          & 35.8          & \multicolumn{1}{l|}{29.4}                                  & 42.5          & 48.8          & 42.9          & 42.5          & 48.8          & \underline{42.9}          \\ 
GLoRIA-ViT  &   MIMIC-CXR   & 224         & 42.0          & 40.9          & \multicolumn{1}{l|}{33.8}                                  & 50.0          & 47.0          & 42.3          & \underline{51.1}          & \underline{49.4}          & 40.8          \\ 
{\color{gray} GLoRIA}   &   {\color{gray} CheXpert*}   & {\color{gray} 224}         & {\color{gray}48.8}          & {\color{gray}46.3}         & \multicolumn{1}{l|}{\color{gray}40.1}                                  & {\color{gray}51.0}          & {\color{gray}49.3}          & {\color{gray}43.5}          & {\color{gray}47.2}          & {\color{gray}46.3}          & {\color{gray}41.5}          \\
{\color{gray}GLoRIA (G+L)}  & {\color{gray}CheXpert*}  & {\color{gray}224}         & {\color{gray}-}             & {\color{gray}-}             & \multicolumn{1}{l|}{{\color{gray}-}}                                     & {\color{gray}50.5}          & {\color{gray}49.0}          & {\color{gray}43.3}          & {\color{gray}47.2}          & {\color{gray}46.3}          & {\color{gray}41.6}         \\ \bottomrule
\end{tabular}
}
\end{table*}
\begin{figure*}[!ht]
  \centering
  \includegraphics[width=\textwidth]{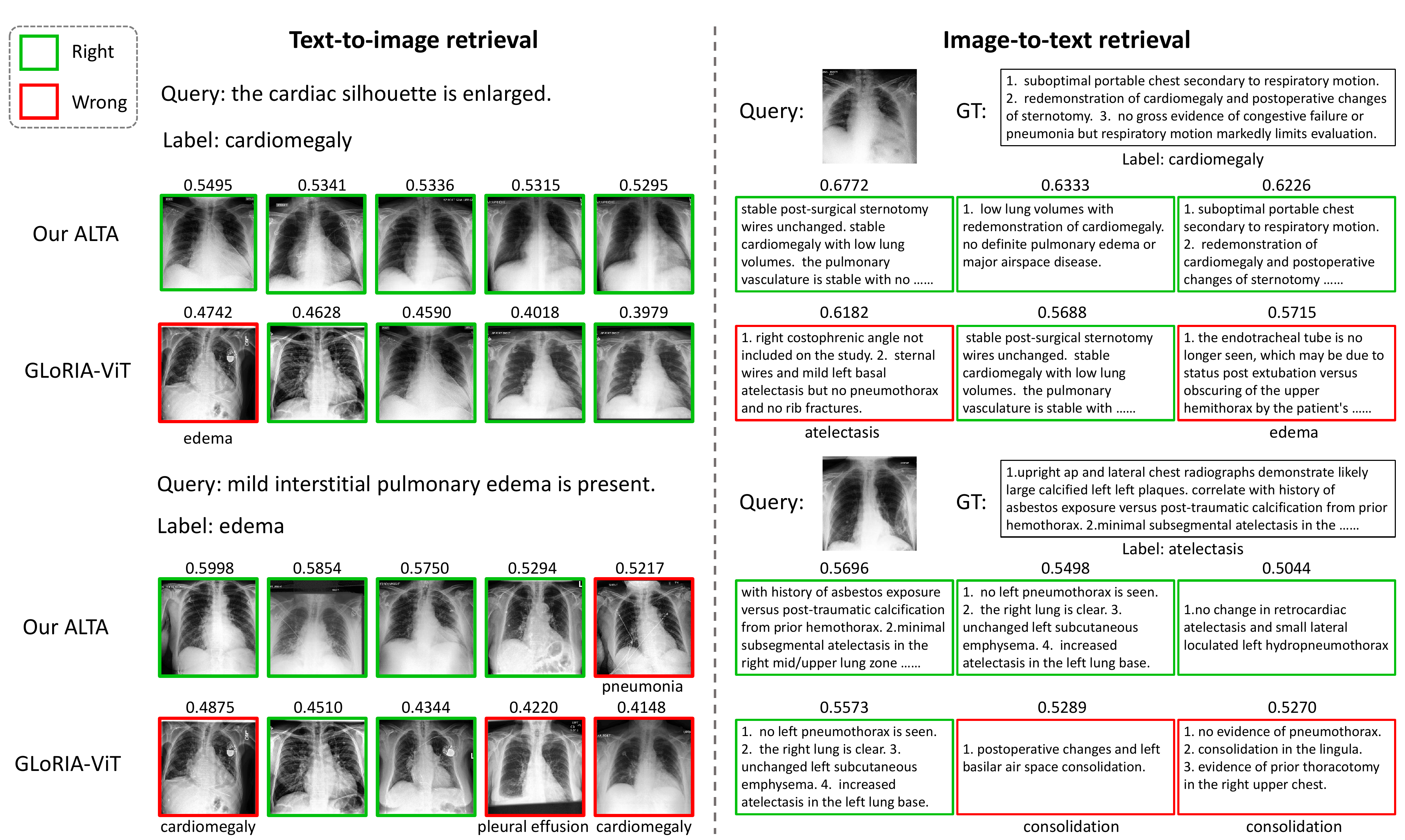}
  \caption{
  Qualitative results of text-to-image and image-to-text retrieval tasks.
  We represent the top five predicted images and the top three texts for ALTA and GLoRIA-ViT, where prediction scores are labeled above the examples.
  We note the categories below wrongly retrieved samples.
  }
  \label{fig: visualization}
\end{figure*}

\textbf{RadNLI} is another language dataset but for non-temporal sentence similarity classification.
To be consistent with MS-CXR-T~\cite{bannur2023learning}, we only retain two categories of paraphrase or contradiction, resulting in 200 text pairs.
We adopt AUC scores to evaluate the performance of zero-shot binary classification.
\subsection{Baselines}
For retrieval tasks, we benchmark the proposed ALTA against four methods based on radiograph-report contrastive learning: ConVIRT~\cite{zhang2022contrastive}, GLoRIA~\cite{huang2021gloria}, BioViL~\cite{boecking2022making}, and BioVil-T~\cite{bannur2023learning}.
ConVIRT~\cite{zhang2022contrastive} learns aligned vision and language representation by contrasting paired radiographs and corresponding radiology reports.
GLoRIA~\cite{huang2021gloria} extends this by incorporating local alignment, contrasting radiograph sub-regions and words from reports.
The evaluations of GLoRIA incorporating global and local representations in vision-language matching tasks are denoted as GLoRIA (G+L).
Since GLoRIA is pretrained using the CheXpert dataset, it has unfair advantages on CheXpert~8$\times$200 and CheXpert~5$\times$200 datasets that are derived from the same dataset.
Relevant results are only presented for reference.
For better comparisons, we reproduce GLoRIA on MIMIC-CXR using ViT-B/16 as the image encoder, denoted as GLoRIA-ViT.
BioViL~\cite{boecking2022making} improves the language model and incorporates masked language modeling during vision-language pretraining, based on which BioViL-T~\cite{bannur2023learning} further adds temporal input to enhance pretraining.
In addition, we also report precision scores under random guess, and the performance of ImageNet pretrained models in the image-to-image retrieval task.

\begin{table}[tb]
\caption{
Results for binary and multi-class zero-shot classification tasks, in ACC, F1, and AUC scores (\%).
The \textbf{best} results are in bold, \underline{second-best} results are underlined.
CX is for CheXpert and $\dagger$ denotes the results cited from the original papers.
* marks the results of the methods with unfair advantages in the CheXpert-based benchmarks.
}
\label{table:zero-shot}
\centering
\resizebox{\linewidth}{!}{
\begin{tabular}{lc|ccc|c}
\toprule
\multicolumn{2}{l|}{Dataset} & \multicolumn{3}{c|}{RSNA Pneumonia}        & CX 5$\times$200             \\ \hline
Method        & Input Size  &ACC~         & F1          & AUC        & ACC                       \\ \hline
\rowcolor{colorlightblue} 
Our ALTA     & 224         & \textbf{83.2}         & \textbf{77.9}        & \textbf{91.0}         & \textbf{56.6}                       \\ \hline
BioVIL$\dagger$       & 480         & 73.2         & 66.5        & 83.1         & -                          \\
BioVIL        & 480         & 76.0         & 73.8        & 86.3         & 43.3                       \\
MedKLIP$\dagger$       & 224         & 80.0         & 63.4        & 86.9         & -                          \\
BioVIL-T$\dagger$     & 448         & 80.5         & 70.6        & 87.1         & -                          \\
BioVIL-T      & 448         & \underline{80.7}         & \underline{76.3}        & 
\underline{89.3}         & 45.7                       \\
GLoRIA-ViT    & 224         & \underline{80.7}         & 75.8        & 88.7         & \underline{47.2}                       \\
GLoRIA        & 224         & 74.2         & 72.4        & 82.4         & \color{gray}{54.9*}                      \\
GLoRIA (G+L)  & 224         & 76.1         & 73.1        & 85.2         & \color{gray}{54.9*}                       \\
\bottomrule
\end{tabular}
}
\end{table}
Besides the methods above, MedKLIP~\cite{wu2023medklip} that enhances radiology vision-language representation by other knowledge information is also compared for zero-shot classification tasks.
Fine-tuning performance on NIH ChestX-ray is applied to evaluate the representation capability of the vision encoder, hence we primarily compare ALTA against MRM, which has been validated to outperform previous report- and self-supervised methods.
For better comparison, results based on ImageNet pretraining, MedKLIP, self-supervised method Model Genesis~\cite{zhou2021models} that includes multiple pretext tasks for representation learning, and REFERS~\cite{zhou2022generalized} that fuses multiview radiograph representation for report-supervised learning are presented as well.
In the temporal image classification task, besides MRM, BioViL, and BioViL-T, we also include CheXRelNet~\cite{karwande2022chexrelnet} that focuses on longitudinal relationships between Chest X-rays and a simple CNN + Transformer model as baselines.
Finally, GLoRIA, BioViL, and BioViL-T are adopted as baselines in sentence similarity classification tasks.
\subsection{Comparative Experiments}
\noindent\textbf{Retrieval tasks.}
To comprehensively assess the performance of the proposed ALTA for vision-language alignment, we design comparative experiments across three retrieval tasks: image-to-image and text-to-image retrieval conducted on CheXpert~8$\times$200, as well as image-to-text retrieval conducted on CheXpert~5$\times$200.
As shown in Table~\ref{table:retrieval}, ALTA consistently delivers substantial improvements across all three retrieval tasks, particularly in cross-modal tasks.
The consistent advancements are statistically significant over baselines at the $p<0.01$ level using a two-sided paired permutation test with 1,000 permutations.
For instance, in the text-to-image retrieval task, ALTA surpasses the best-performing baseline method by 5.0\%, 8.5\%, and 6.5\% in terms of absolute points of P@5, P@10, and P@50, respectively.

In the image-to-text retrieval tasks, our ALTA outperforms GLoRIA-ViT, which is the best among the baselines, by 5.7\%, 6.0\%, and 6.0\% in absolute points of P@5, P@10, and P@100, respectively.
In the image-to-image retrieval task, we observe that MRM underperforms most other comparative methods, showing that the absence of explicit vision-language alignment hinders intra-modal retrieval tasks as well.
In contrast, our approach aligns the vision representations of MRM to language representations, resulting in substantial improvements in the image-to-image retrieval task.
Specifically, our ALTA outperforms the best baseline by 6.0\%, 4.1\%, and 4.4\% in terms of absolute points of P@5, P@10, and P@50, respectively.
For more intuitive analyses, we present visualization results comparing GLoRIA-ViT~\cite{huang2021gloria} and our ALTA in text-to-image and image-to-text retrieval tasks in Fig.~\ref{fig: visualization}, showing that our ALTA achieves superior accuracy and higher prediction scores.
These quantitative and qualitative analyses indicate that the proposed ALTA enhances the performance of both intra-modal and inter-modal retrieval.

\noindent\textbf{Zero-shot classification tasks.}
For a more comprehensive evaluation of vision-language alignment performance, we conduct comparative zero-shot classification experiments on the RSNA Pneumonia and CheXpert~5$\times$200 datasets, which are analyzed quantitatively in Table~\ref{table:zero-shot}.
MRM is not included in this comparison because it does not explicitly align vision and language representations, making it inapplicable for direct application in zero-shot classification.
The experimental results indicate that, despite underperformance in retrieval tasks, BioViL-T emerges as the leading baseline method in binary classification evaluation conducted on RNSA Pneumonia.
The proposed ALTA outperforms it by 2.5\% and 1.6\% in terms of absolute accuracy and F1 score, respectively.
Notably, even with an AUC score approaching approximately 90\%, our method still shows a 1.7\% absolute improvement, achieving 91.0\%.
In the multi-class zero-shot classification task on CheXpert~5$\times$200, GLoRIA performs best among the baseline methods and our ALTA achieves a 1.7\% absolute improvement in accuracy.
By utilizing a two-sided paired permutation test with 1,000 permutations, we observe statistically significant improvements ($p<0.01$) of our method.
\begin{table}[!b]
\caption{Results on NIH ChestX-ray.
AUC scores (\%) are displayed.
The \textbf{best} results are in bold, and \underline{second-best} results are underlined.
* denotes results using LoRA fine-tuning.
}
\label{table:NIH}
\centering
\resizebox{0.9\linewidth}{!}{
\renewcommand\arraystretch{1}
\begin{tabular}{lccc}
\toprule
\multicolumn{1}{c|}{}                                     & \multicolumn{3}{c}{Labeling ratio} \\ \cline{2-4} 
\multicolumn{1}{l|}{\multirow{-2}{*}{Method}}             & ~1\%~       & ~10\%~      & ~100\%~      \\ \hline
\rowcolor{colorlightblue} 
\multicolumn{1}{l|}{\cellcolor{colorlightblue}Our   ALTA} & \textbf{80.7}{\scriptsize$\pm$0.2}      & \textbf{84.3}{\scriptsize$\pm$0.1}      & \textbf{86.0}{\scriptsize$\pm$0.1}       \\ \hline
\multicolumn{1}{l|}{MRM*}                                & 80.6{\scriptsize$\pm$0.2}      & 84.0{\scriptsize$\pm$0.2}      & 85.8{\scriptsize$\pm$0.1}       \\
\multicolumn{1}{l|}{MRM}                                 & 79.4{\scriptsize$\pm$0.8}      & 84.0{\scriptsize$\pm$0.5}      & 85.9{\scriptsize$\pm$0.3}       \\ \hline
\multicolumn{1}{l|}{REFERS}                              & 76.7      & 80.9      & 84.7       \\
\multicolumn{1}{l|}{MedKLIP}                             & 77.2      & 78.9      & 83.2       \\
\multicolumn{1}{l|}{Model genesis}                     & 70.3      & 76.0      & 81.0       \\
\multicolumn{1}{l|}{ImageNet pretraining~~}             & 69.8      & 74.4      & 80.1       \\ \bottomrule
\end{tabular}
}
\end{table}
\begin{table}[!t]
\caption{
Results of temporal image classification on the MS-CXR-T dataset.
Macro accuracies (\%) are displayed.
TF is short for transformer, Pl. effusion and PTX denotes pleural effusion and pneumothorax, respectively.
The \textbf{best} results are in bold, while the \underline{second-best} results are underlined.
}
\label{table:temporal_image}
\resizebox{\linewidth}{!}{
\begin{tabular}{cccccc}
\toprule
Method            & \textbf{Consolidation} & \textbf{Pl. effusion} & \textbf{Pneumonia} & \textbf{PTX} & \textbf{Edema}    \\ \hline
\rowcolor{colorlightblue} 
Our ALTA          & \textbf{62.2±1.6}      & \textbf{69.0±0.6}         & \textbf{62.4±0.9}  & \textbf{46.6±1.1}     & \textbf{69.5±0.6} \\ \hline
MRM               & 58.9±2.2               & 62.1±1.4                  & 61.1±1.3           & 41.5±0.9              & 68.0±0.9          \\
BioViL-T          & \underline{61.1±2.4}               & \underline{67.0±0.8}                  & \underline{61.9±1.9}           & \underline{42.6±1.6}              & \underline{68.5±0.8}          \\
BioViL            & 56.1±1.5               & 62.3±1.1                  & 59.4±1.0           & 41.7±2.8              & 67.5±0.8          \\
CheXRelNet        & 47                     & 47                        & 47                 & 36                    & 49                \\
CNN + TF & 44.0±2.0               & 61.3±1.6                  & 45.1±3.5           & 31.5±3.1              & 65.5±1.1          \\
\bottomrule
\end{tabular}
}
\end{table}

\noindent\textbf{Evaluations of vision understanding.}
Table~\ref{table:NIH} validates the radiograph understanding capability of the vision encoder after implementing the proposed efficient alignment.
Since LoRA can enhance fine-tuning for downstream tasks~\cite{lian2024less}, we conduct LoRA fine-tuning and report the results of MRM based on LoRA fine-tuning for fair comparisons.
Experiments in NIH ChestX-ray show that the proposed ALTA exhibits comparative performances with MRM and outperforms other self-supervised and report-supervised methods.
The results reflect that the proposed efficient alignment through adaptation does not impede the capability to vision understanding, which is crucial for an adaptation method.

Beyond the non-temporal radiograph classification task, the ability of temporal vision understanding is evaluated on the MS-CXR-T dataset by temporal image classification.
As shown in Table~\ref{table:temporal_image}, our ALTA exhibits consistent advantages over all baseline methods.
Benefiting from pertaining with temporal images, BioViL emerges as the top-performing baseline method.
However, the proposed approach outperforms BioViL, achieving notable improvements for some abnormalities such as pneumothorax.
The comparative experiments for temporal and non-temporal image understanding tasks validate the vision comprehension capability of the proposed ALTA.

\begin{table}[!b]
\caption{Results for sentence similarity classification tasks in terms of ACC and AUC (\%).
The \textbf{highest} scores apart from ablation analysis are highlighted in bold, and the \underline{second-best} results are underlined.
}
\label{table:language}
\centering
\resizebox{\linewidth}{!}{
\renewcommand\arraystretch{0.96}
\begin{tabular}{lcccc}
\toprule
\multicolumn{1}{l|}{}                              & \multicolumn{2}{c|}{MS-CXR-T}        & \multicolumn{2}{c}{RadNLI} \\ \cline{2-5} 
\multicolumn{1}{l|}{\multirow{-2}{*}{Method}}      & ~~~~~~~ACC~~~~~   & \multicolumn{1}{c|}{~~~~~AUC~~~~~~~}                        & ~~~~~~~ACC~~~~~                     & ~~~~~AUC~~~~~~~                 \\ \hline
\multicolumn{1}{l|}{GLoRIA}                        & 39.05 & \multicolumn{1}{c|}{52.58}                         & 54.00                   & 67.73                  \\
\multicolumn{1}{l|}{GLoRIA-ViT}                       & 39.05 & \multicolumn{1}{c|}{55.70}                         & 68.00                   & 81.59                  \\
\multicolumn{1}{l|}{BioViL}                        & 76.74 & \multicolumn{1}{c|}{83.74}                         & \underline{81.50}                   & \underline{87.21}                  \\
\multicolumn{1}{l|}{BioViL-T}                      & \underline{86.71} & \multicolumn{1}{c|}{\underline{92.85}}                         & 76.00                   & 82.39                  \\ \hline
\rowcolor{colorlightblue} 
\multicolumn{1}{l|}{\cellcolor{colorlightblue}Our ALTA} & \textbf{87.54} & \multicolumn{1}{c|}{\cellcolor{colorlightblue}\textbf{93.80}} & \textbf{82.50}                   & \textbf{89.88}                  \\ \hline
\textit{Ablation analysis}                         &       &                                                    &                         &                        \\
\multicolumn{1}{l|}{{\color{red} -} Temporal}                    & 81.46 (\color{red} $-$6.08) & \multicolumn{1}{c|}{88.29 (\color{red} $-$5.51)}                         & 81.50 (\color{red} $-$1.00)                  & 87.21 (\color{red} $-$2.67)                 \\
\multicolumn{1}{l|}{{\color{red} -} Multiview}                  & 86.16 (\color{red} $-$1.38) & \multicolumn{1}{c|}{91.88 (\color{red} $-$1.92)}                         & 80.50 (\color{red} $-$2.00)                  & 89.22 (\color{red} $-$0.66)                 \\
\multicolumn{1}{l|}{{\color{red} -} Temporal \& Multiview}       & 80.08 (\color{red} $-$7.46) & \multicolumn{1}{c|}{87.54 (\color{red} $-$6.26)}                         & 81.50 (\color{red} $-$1.00)                  & 89.26 (\color{red} $-$0.62)                 \\
\multicolumn{1}{l|}{{\color{red} -} Local Loss}                  & 87.55 (\color{green} $+$0.01) & \multicolumn{1}{c|}{94.17 (\color{green} $+$0.37)} & 79.00 (\color{red} $-$3.50)                 & 87.66 (\color{red} $-$2.22)                 \\
\multicolumn{1}{l|}{{\color{red} -} MLM Loss}                    & 84.50 (\color{red} $-$3.04) & \multicolumn{1}{c|}{92.00 (\color{red} $-$1.80)}                         & 86.00 (\color{green} $+$3.50)      & 89.40 (\color{red} $-$0.48)                 \\
\multicolumn{1}{l|}{{\color{red} -} MIM Loss}                    & 87.56 (\color{green} $+$0.02) & \multicolumn{1}{c|}{94.92 (\color{green} $+$1.12)} & 81.00 (\color{red} $-$1.50)                  & 87.92 (\color{red} $-$1.96)                 \\ \bottomrule
\end{tabular}
}
\end{table}
\begin{table*}[!th]
\caption{Ablation analysis for retrieval and zero-shot classification (abbreviated as ZS cls.) tasks.
}
\label{table:ablation}
\centering
\resizebox{\linewidth}{!}{
\begin{tabular}{lccccccccccccc}
\toprule
\multicolumn{1}{l|}{Dataset}                                    & \multicolumn{6}{c|}{CheXpert 8$\times$200}                                                                                                                                                                                                     & \multicolumn{4}{c|}{CheXpert 5$\times$200}                                                                                                                                     & \multicolumn{3}{c}{RSNA Pneumonia}                                                           \\ \hline
\multicolumn{1}{l|}{Task}                                       & \multicolumn{3}{c|}{Image $\rightarrow$ Image}                                                                                   & \multicolumn{3}{c|}{Text $\rightarrow$ Image}                                                                                    & \multicolumn{3}{c|}{Image $\rightarrow$ Text}                                                                                    & \multicolumn{1}{c|}{ZS  cls.}                      & \multicolumn{3}{c}{ZS   cls.}                                                                 \\ \hline
\multicolumn{1}{l|}{Method}                                     & P@5                           & P@10                          & \multicolumn{1}{c|}{P@50}                          & P@5                           & P@10                          & \multicolumn{1}{c|}{P@50}                          & P@5                           & P@10                          & \multicolumn{1}{c|}{P@100}                         & \multicolumn{1}{c|}{ACC}                          & ACC                          & F1                            & AUC                         \\ \hline
\rowcolor{colorlightblue} 
\multicolumn{1}{l|}{\cellcolor{colorlightblue}Our ALTA}              & 51.3                          & 47.1                          & \multicolumn{1}{c|}{\cellcolor{colorlightblue}38.7}  & 64.5                          & 65.8                          & \multicolumn{1}{c|}{\cellcolor{colorlightblue}53.0}  & 56.8                          & 55.4                          & \multicolumn{1}{c|}{\cellcolor{colorlightblue}46.8}  & \multicolumn{1}{c|}{\cellcolor{colorlightblue}56.6}  & 83.2                          & 77.9                          & 91.0                          \\ \hline
\multicolumn{2}{l}{\textit{Ablation study of radiograph inputs}}                                                                  &                               &                                                    &                               &                               &                                                    &                               &                               &                                                    &                                                    &                               &                               &                               \\
\multicolumn{1}{l|}{}                                           & 45.5                          & 44.8                          & \multicolumn{1}{c|}{36.4}                          & 62.0                          & 61.0                          & \multicolumn{1}{c|}{50.9}                          & 55.2                          & 54.3                          & \multicolumn{1}{c|}{45.9}                          & \multicolumn{1}{c|}{55.5}                          & 81.7                          & 76.3                          & 91.0                          \\
\multicolumn{1}{l|}{\multirow{-2}{*}{{\color{red} -} Temporal}}             & {\color{red} ($-$5.8)} & {\color{red} ($-$2.3)} & \multicolumn{1}{c|}{{\color{red} ($-$2.3)}} & {\color{red} ($-$2.5)} & {\color{red} ($-$4.8)} & \multicolumn{1}{c|}{{\color{red} ($-$2.1)}} & {\color{red} ($-$1.6)} & {\color{red} ($-$1.1)} & \multicolumn{1}{c|}{{\color{red} ($-$0.9)}} & \multicolumn{1}{c|}{{\color{red} ($-$1.1)}} & {\color{red} ($-$1.5)} & {\color{red} ($-$1.6)} & { (0.0)}  \\ \hline
\multicolumn{1}{l|}{}                                           & 48.0                          & 44.1                          & \multicolumn{1}{c|}{35.8}                          & 68.0                          & 62.3                          & \multicolumn{1}{c|}{51.7}                          & 54.8                          & 54.0                          & \multicolumn{1}{c|}{45.6}                          & \multicolumn{1}{c|}{56.8}                          & 80.3                          & 76.3                          & 90.0                          \\
\multicolumn{1}{l|}{\multirow{-2}{*}{{\color{red} -} Multiview}}            & {\color{red} ($-$3.3)} & {\color{red} ($-$3.0)} & \multicolumn{1}{c|}{{\color{red} ($-$2.9)}} & {\color{green} (+3.5)} & {\color{red} ($-$3.5)} & \multicolumn{1}{c|}{{\color{red} ($-$1.3)}} & {\color{red} ($-$2.0)} & {\color{red} ($-$1.4)} & \multicolumn{1}{c|}{{\color{red} ($-$1.2)}} & \multicolumn{1}{c|}{{\color{green} (+0.2)}} & {\color{red} ($-$2.9)} & {\color{red} ($-$1.6)} & {\color{red} ($-$1.0)} \\ \hline
\multicolumn{1}{l|}{}                                           & 44.5                          & 41.5                         & \multicolumn{1}{c|}{34.8}                          & 60.0                          & 59.3                          & \multicolumn{1}{c|}{51.2}                          & 55.5                          & 54.3                          & \multicolumn{1}{c|}{45.3}                          & \multicolumn{1}{c|}{55.2}                          & 81.5                          & 76.5                          & 90.3                          \\
\multicolumn{1}{l|}{\multirow{-2}{*}{{\color{red} -} Temporal \& Multiview}} & {\color{red} ($-$6.8)} & {\color{red} ($-$5.6)} & \multicolumn{1}{c|}{{\color{red} ($-$3.9)}} & {\color{red} ($-$4.5)} & {\color{red} ($-$6.5)} & \multicolumn{1}{c|}{{\color{red} ($-$1.8)}} & {\color{red} ($-$1.3)} & {\color{red} ($-$1.1)} & \multicolumn{1}{c|}{{\color{red} ($-$1.5)}} & \multicolumn{1}{c|}{{\color{red} ($-$1.4)}} & {\color{red} ($-$1.7)} & {\color{red} ($-$1.4)} & {\color{red} ($-$0.7)} \\ \hline
\multicolumn{1}{l|}{}                                           & 45.5                          & 44.3                         & \multicolumn{1}{c|}{36.6}                          & 61.5                          & 58.8                          & \multicolumn{1}{c|}{50.6}                          & 51.2                          & 50.6                          & \multicolumn{1}{c|}{44.5}                          & \multicolumn{1}{c|}{55.0}                          & 79.7                          & 75.1                          & 90.1                          \\
\multicolumn{1}{l|}{\multirow{-2}{*}{\makecell[c]{{\color{red} -} Temporal \& Multiview \\(including lateral images) }}} & {\color{red} ($-$5.8)} & {\color{red} ($-$2.8)} & \multicolumn{1}{c|}{{\color{red} ($-$2.1)}} & {\color{red} ($-$3.0)} & {\color{red} ($-$7.0)} & \multicolumn{1}{c|}{{\color{red} ($-$2.4)}} & {\color{red} ($-$5.6)} & {\color{red} ($-$4.8)} & \multicolumn{1}{c|}{{\color{red} ($-$2.3)}} & \multicolumn{1}{c|}{{\color{red} ($-$1.6)}} & {\color{red} ($-$3.5)} & {\color{red} ($-$2.8)} & {\color{red} ($-$0.9)}
\\ \hline
\multicolumn{2}{l}{\textit{Ablation study of loss functions}}                                                            &                               &                                                    &                               &                               &                                                    &                               &                               &                                                    &                                                    &                               &                               &                               \\
\multicolumn{1}{l|}{}                                           & 48.3                          & 46.1                          & \multicolumn{1}{c|}{36.5}                          & 67.0                          & 63.8                          & \multicolumn{1}{c|}{55.8}                          & 56.9                          & 55.9                          & \multicolumn{1}{c|}{46.7}                          & \multicolumn{1}{c|}{54.4}                          & 78.9                          & 75.3                          & 89.3                          \\
\multicolumn{1}{l|}{\multirow{-2}{*}{{\color{red} -} Local Loss}}             & {\color{red} ($-$3.0)} & {\color{red} ($-$1.0)} & \multicolumn{1}{c|}{{\color{red} ($-$2.2)}} & {\color{green} (+2.5)} & {\color{red} ($-$2.0)} & \multicolumn{1}{c|}{{\color{green} (+2.8)}} & {\color{green} (+0.1)} & {\color{green} (+0.5)} & \multicolumn{1}{c|}{{\color{red} ($-$0.1)}} & \multicolumn{1}{c|}{{\color{red} ($-$2.2)}} & {\color{red} ($-$4.3)} & {\color{red} ($-$2.6)} & {\color{red} ($-$1.7)} \\ \hline
\multicolumn{1}{l|}{}                                           & 44.3                          & 44.9                          & \multicolumn{1}{c|}{37.8}                          & 68.0                          & 60.0                          & \multicolumn{1}{c|}{50.9}                          & 55.6                          & 54.3                          & \multicolumn{1}{c|}{46.7}                          & \multicolumn{1}{c|}{56.4}                          & 79.3                          & 74.2                          & 88.3                          \\
\multicolumn{1}{l|}{\multirow{-2}{*}{{\color{red} -} MLM Loss}}               & {\color{red} ($-$7.0)} & {\color{red} ($-$2.2)} & \multicolumn{1}{c|}{{\color{red} ($-$0.9)}} & {\color{green} (+3.5)} & {\color{red} ($-$5.8)} & \multicolumn{1}{c|}{{\color{red} ($-$2.1)}} & {\color{red} ($-$1.2)} & {\color{red} ($-$1.1)} & \multicolumn{1}{c|}{{\color{red} ($-$0.1)}} & \multicolumn{1}{c|}{{\color{red} ($-$0.2)}} & {\color{red} ($-$3.9)} & {\color{red} ($-$3.7)} & {\color{red} ($-$2.7)} \\ \hline
\multicolumn{1}{l|}{}                                           & 49.5                          & 46.8                          & \multicolumn{1}{c|}{37.1}                          & 64.5                          & 62.0                          & \multicolumn{1}{c|}{50.9}                          & 56.1                          & 54.5                          & \multicolumn{1}{c|}{47.0}                          & \multicolumn{1}{c|}{52.8}                          & 78.9                          & 75.3                          & 89.8                          \\
\multicolumn{1}{l|}{\multirow{-2}{*}{{\color{red} -} MIM Loss}}               & {\color{red} ($-$1.8)} & {\color{red} ($-$0.3)} & \multicolumn{1}{c|}{{\color{red} ($-$1.6)}} & { (0.0)}  & {\color{red} ($-$3.8)} & \multicolumn{1}{c|}{{\color{red} ($-$2.1)}} & {\color{red} ($-$0.7)} & {\color{red} ($-$0.9)} & \multicolumn{1}{c|}{{\color{green} (+0.2)}} & \multicolumn{1}{c|}{{\color{red} ($-$3.8)}} & {\color{red} ($-$4.3)} & {\color{red} ($-$2.6)} & {\color{red} ($-$1.2)} \\ \bottomrule
\end{tabular}
}
\end{table*}
\noindent\textbf{Language understanding analysis.}
The comparative experiments in temporal (on MS-CXR-T) and non-temporal (on RadNLI) zero-shot sentence similarity classification tasks validate the enhancement of ALTA in language understanding.
As Table~\ref{table:language} shows, our ALTA outperforms all baseline methods on sentence similarity classification tasks.
Benefiting from the incorporation of temporal radiograph inputs, BioViL-T shows observable improvement over BioViL in the temporal sentence similarity classification task.
However, its performance decreases notably for non-temporal sentence similarity classification.
In contrast, the proposed ALTA exhibits robust performance for both temporal and non-temporal tasks.
Specifically, ALTA surpasses BioViL in the temporal sentence similarity classification task thanks to the temporal alignment design and is comparative to BioViL-T.
Meanwhile, ALTA outperforms BioViL-T in the non-temporal sentence similarity classification task and is on par with BioViL.
The comprehensive experiments validate the capability of the proposed ALTA for temporal and non-temporal language understanding.
\subsection{Ablation Analysis of Temporal-Multiview Inputs}
To verify the effectiveness and necessity of temporal-multiview radiograph inputs, we design comprehensive ablation experiments on temporal and multiview inputs.
The performances are evaluated in three retrieval tasks, two zero-shot classification tasks, and two language understanding tasks.

\noindent\textbf{The necessity of temporal radiograph inputs.}
When temporal radiograph inputs are not included, i.e., the radiograph inputs are changed from the image quaternion $(\mathcal{I}_{cf}, \mathcal{I}_{c\ell}, \mathcal{I}_{pf}, \mathcal{I}_{p\ell})$ to image pair $(\mathcal{I}_{cf}, \mathcal{I}_{c\ell})$, notable degradations of performance can be observed in all retrieval, zero-shot classification, and language understanding tasks, as shown in Table~\ref{table:language} and~\ref{table:ablation}.
In particular, the performance drops severely on the temporal sentence similarity classification task, suggesting that the temporal radiograph inputs contribute notably to temporal sentence understanding.

\noindent\textbf{The benefits of multiview radiograph inputs.}
When multiview inputs are not included, i.e., the radiograph inputs are reduced to the image pair $(\mathcal{I}_{cf}, \mathcal{I}_{pf})$, where the performance also decreases across all tasks, although to a slightly less extent compared to the case of removing temporal inputs, showing the benefits of introducing multiview radiographs.

\noindent\textbf{The impact of removing temporal and multiview inputs concurrently.}
In the most extreme scenario, where both temporal and multiview inputs are removed and only the current frontal radiograph $\mathcal{I}_{cf}$ is included, the model suffers the most severe performance degradation, which is evidenced by decreases in all metrics across each task.
This indicates that including images from the prior study and lateral view facilitates better vision-language alignment.
In addition, to alleviate the ambiguity regarding whether the performance degradation stems from the exclusion of lateral radiographs, we also conduct experiments using all images without employing the proposed temporal-view designs.
The results presented in Table~\ref{table:ablation} show that the performances are comparable regardless of whether lateral images are included.
\noindent\subsection{Ablation Analysis of Loss Functions}
\label{subsec: ablation_of_loss_short}
Our ALTA incorporates four loss functions.
Apart from the mandatory global alignment loss~\eqref{eq:global_loss}, we conduct ablation analysis by removing each of the other three loss functions: local alignment loss, MLM loss, and MIM loss.
We evaluate their impact on retrieval, zero-shot classification, and language understanding tasks.
Table~\ref{table:language} reveals that for the sentence similarity zero-shot classification task, the performance decreases to varying degrees when any of the loss functions are removed.
As shown in Table~\ref{table:ablation}, the circumstances for retrieval and zero-shot classification are more complex as discussed below:

\begin{table}[!ht]
\caption{
Ablation analysis on masking ratios of input images and reports.
Retrieval results are reported with the \textbf{best} ones in bold and the \underline{second-best} ones underlined.
}
\label{table:masking_ablation}
\centering
\resizebox{\linewidth}{!}{
\begin{tabular}{c|cccccc|ccc}
\toprule
Image              & \multicolumn{6}{c|}{CheXpert 8$\times$200}                                                                                & \multicolumn{3}{c}{CheXpert 5$\times$200}            \\ \cline{2-10}
masking           & \multicolumn{3}{c|}{Image$\rightarrow$Image}                                   & \multicolumn{3}{c|}{Text$\rightarrow$Image}               & \multicolumn{3}{c}{Image$\rightarrow$Text}                \\ \cline{2-10}
         ratio                     & P@5           & P@10          & \multicolumn{1}{c|}{P@50}          & P@5           & P@10          & P@50          & P@5           & P@10          & P@100         \\ \hline
50\%                          & \underline{49.0}          & \underline{46.8}          & \multicolumn{1}{c|}{\underline{37.4}}          & \textbf{67.0} & 65.0          & \textbf{54.6} & \underline{55.3}          & \underline{54.6}          & \textbf{46.8} \\
\rowcolor{colorlightblue}
75\%                          & \textbf{51.3} & \textbf{47.1} & \multicolumn{1}{c|}{\textbf{38.7}} & \underline{64.5}          & \textbf{65.8} & \underline{53.0}          & \textbf{56.8} & \textbf{55.4} & \textbf{46.8} \\
85\%                          & 46.8          & 44.6          & \multicolumn{1}{c|}{\underline{37.4}}          & 62.5          & 61.6          & 52.5          & 54.9          & 52.2          & 43.9          \\ \hline \hline
Report              & \multicolumn{6}{c|}{CheXpert 8×200}                                                                                & \multicolumn{3}{c}{CheXpert 5×200}            \\ \cline{2-10}
masking & \multicolumn{3}{c|}{Image$\rightarrow$Image}                                   & \multicolumn{3}{c|}{Text$\rightarrow$Image}               & \multicolumn{3}{c}{Image$\rightarrow$Text}                \\ \cline{2-10}
       ratio                       & P@5           & P@10          & \multicolumn{1}{c|}{P@50}          & P@5           & P@10          & P@50          & P@5           & P@10          & P@100         \\ \hline
25\%                          & \underline{50.5}          & \underline{45.1}          & \multicolumn{1}{c|}{\textbf{39.3}} & \underline{62.4}          & \underline{61.3}          & \underline{52.9}          & \textbf{58.3} & \underline{54.2}          & \underline{45.5}          \\
\rowcolor{colorlightblue}
50\%                          & \textbf{51.3} & \textbf{47.1} & \multicolumn{1}{c|}{\underline{38.7}}          & \textbf{64.5} & \textbf{65.8} & \textbf{53.0} & \underline{56.8}          & \textbf{55.4} & \textbf{46.8} \\
75\%                          & 45.8          & 44.6          & \multicolumn{1}{c|}{37.8}          & 60.2          & 60.1          & 50.5          & 55.3          & 52.7          & 44.6          \\ \bottomrule
\end{tabular}
}
\end{table}
\textbf{The local alignment loss~\eqref{eq:local_loss}~}plays an important role in image-to-image retrieval and zero-shot classification tasks, as its removal leads to considerable performance degradation.
Although no noticeable performance degradation is observed for retrieval tasks between images and text, we retain the local loss function as the default setting of our ALTA based on comprehensive considerations.

\textbf{When the MLM loss~\eqref{eq:MLM_loss}~}is removed, there is a notable impact on performance.
Specifically, the P@5 metric in the image-to-image retrieval task experiences a 7.0\% decrease.
In zero-shot binary classification, the accuracy, F1 score, and AUC score decrease by 3.9\%, 3.7\%, and 2.7\%, respectively.

\textbf{The removal of the MIM loss~\eqref{eq:MIM_loss}~}leads to degradation of varying degrees across almost all metrics of the five retrieval and zero-shot classification tasks.
Notably, accuracy decreases by 3.8\% for the multi-class zero-shot classification task, which is the most significant change for the task among the ablation experiments mentioned above.

In conclusion, the ablation analyses underscore the indispensability and efficacy of these loss functions.
Therefore, we incorporate all of them in our ALTA.
\subsection{Ablation Analysis of Masking Ratios}
To maintain masked modeling supervision and accelerate training, we conduct random masking on input images and reports.
Here we explain the default settings of the masking ratio through ablation analysis, reported in Table~\ref{table:masking_ablation}.
Experimentally, we set the default image and report masking ratios to 75\% and 50\%, respectively.
\begin{figure}[h]
  \centering
  \resizebox{\linewidth}{!}{
  \includegraphics[width=\textwidth]{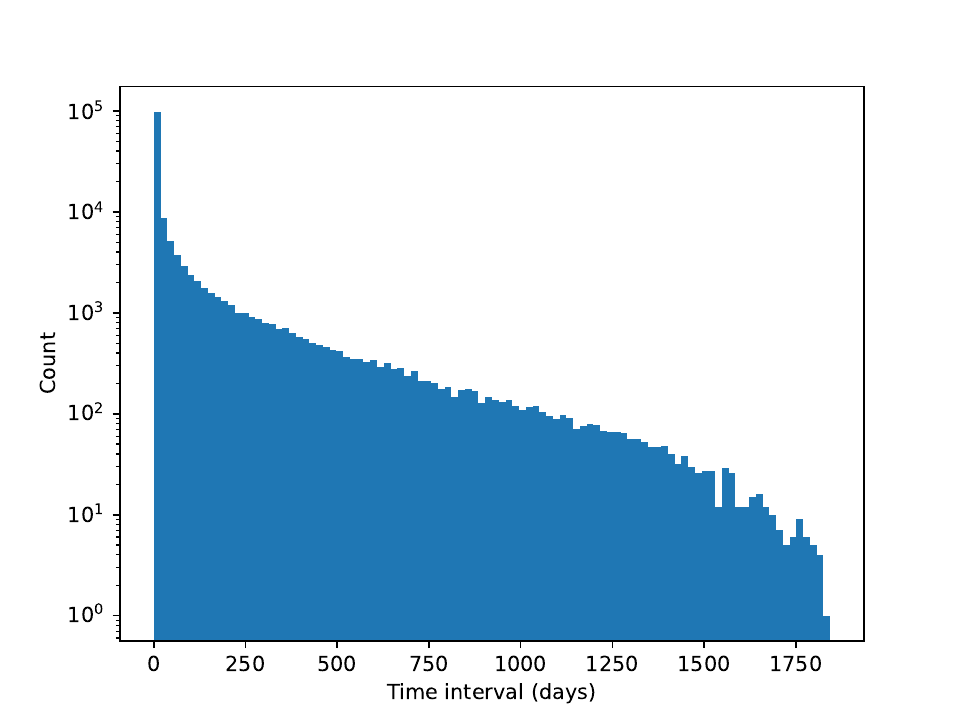}
  }
  \caption{
  Log-scale histogram visualization of the time interval distribution between prior and current studies in the restructured temporal-multiview MIMIC-CXR dataset.
  }
  \label{fig: time_interval_distribution}
\end{figure}
\subsection{Time Interval Distribution Analysis}
To better understand the time interval distribution between prior and current studies in the restructured temporal-multiview MIMIC-CXR dataset, we visualize the distribution using a histogram, as depicted in Fig.~\ref{fig: time_interval_distribution}.
The histogram reveals that the time intervals follow a long-tailed distribution.
Leveraging this statistical insight, we investigate the impact of time distribution on model performance by conducting experiments under three distinct data settings: (1) randomly selected 50\% of the temporal data, (2) the smaller 50\% of the data (shorter time intervals), and (3) the larger 50\% of the data (longer time intervals).
As shown in Table~\ref{table:time_interval_distribution}, the results across these settings are comparable, indicating that our ALTA framework exhibits robustness to variations in time intervals.
The results underscore the applicability of the proposed method in various clinical scenarios with varying time interval distributions.

\begin{table}[!t]
\caption{
Time interval analysis that utilizes randomly selected 50\% of the temporal data, the smaller 50\% of the data, and the larger 50\% of the data to conduct ALTA training and evaluate the performance of retrieval tasks.
}
\label{table:time_interval_distribution}
\centering
\renewcommand{\arraystretch}{1.2}
\resizebox{\linewidth}{!}{
\begin{tabular}{c|cccccc|ccc}
\toprule
\multirow{3}{*}{\makecell[c]{Time interval}} & \multicolumn{6}{c|}{CheXpert 8$\times$200}                                        & \multicolumn{3}{c}{CheXpert 5$\times$200} \\ \cline{2-10} 
                               & \multicolumn{3}{c|}{Image$\rightarrow$Image}        & \multicolumn{3}{c|}{Text$\rightarrow$Image} & \multicolumn{3}{c}{Image$\rightarrow$Text}    \\ \cline{2-10} 
                               & P@5  & P@10 & \multicolumn{1}{c|}{P@50} & P@5       & P@10     & P@50     & P@5       & P@10      & P@100     \\ \hline
Random 50\%                        & 45.6 & 47.3 & \multicolumn{1}{c|}{37.7} & 63.5      & 62.3     & 53.6     & 55.3      & 54.1      & 45.1      \\
Smaller 50\%                & 46.5 & 47.7 & \multicolumn{1}{c|}{38.2} & 64.0      & 64.5     & 51.5     & 53.0      & 52.5      & 44.9      \\
Larger 50\%                 & 46.5 & 43.8 & \multicolumn{1}{c|}{37.1} & 68.0      & 62.0     & 52.4     & 55.5      & 54.2      & 45.8      \\
\bottomrule
\end{tabular}
}
\end{table}
\subsection{Analysis of Parameter-Efficient Training Implementation}
In the proposed ALTA, we implement parameter-efficient training using adapters~\cite{houlsby2019parameter}, which have been widely used for various parameter-efficient training tasks.
Since our framework is agnostic to specific adaptation modules, we evaluate the performance of adapters (the default configuration) against LoRA~\cite{hu2021lora}, another prevalent method for parameter-efficient training.
For a comprehensive comparison, we also present results from full-model training.
As shown in Table~\ref{table:adapter_lora}, both adapter-based and LoRA-based parameter-efficient training outperform full training, while training only 8\% of the parameters.
The performance advantage of adaptation methods over full training is assumed to stem from the mitigation of catastrophic forgetting in pretrained vision and language models during alignment training.
Since adapter-based ALTA slightly outperforms LoRA, we adopt adapters as the default parameter-efficient training modules in our approach.
Notably, ALTA with LoRA also exhibits promising results, indicating the robustness of our vision-language alignment framework across different parameter-efficient mechanisms.
\begin{table}[!h]
\caption{
Analysis of parameter-efficient training implementation by comparing the performance of our ALTA with adapters or LoRA modules.
The results of full training are also included for comprehensive comparison.
}
\label{table:adapter_lora}
\centering
\renewcommand{\arraystretch}{1.2}
\resizebox{\linewidth}{!}{
\begin{tabular}{c|cccccc|ccc}
\toprule
\multirow{3}{*}{\begin{tabular}[c]{@{}c@{}}Training\\ implementation\end{tabular}} & \multicolumn{6}{c|}{CheXpert 8$\times$200}                                & \multicolumn{3}{c}{CheXpert 5$\times$200} \\ \cline{2-10} 
                        & \multicolumn{3}{c|}{Image$\rightarrow$Image} & \multicolumn{3}{c|}{Text$\rightarrow$Image} & \multicolumn{3}{c}{Image$\rightarrow$Text}     \\ \cline{2-10}
                               & P@5  & P@10 & \multicolumn{1}{c|}{P@50} & P@5       & P@10     & P@50     & P@5       & P@10      & P@100     \\ \hline
\rowcolor{colorlightblue}
Adapter training                                                                    & 51.3                 & 47.1                 & \multicolumn{1}{c|}{38.7} & 64.5                 & 65.8                 & \multicolumn{1}{c|}{53.0} & 56.8                 & 55.4                 & 46.8                 \\
LoRA training                                                                       & 49.8                 & 47.5                 & \multicolumn{1}{c|}{38.7} & 61.5                 & 62.0                 & \multicolumn{1}{c|}{51.9} & 54.4                 & 54.1                 & 45.9                 \\                                        Full training                          & 47.5                 & 45.8                 & \multicolumn{1}{c|}{34.9} & 61.0                 & 61.0                 & \multicolumn{1}{c|}{53.5} & 52.5                 & 50.8                 & 42.3                 \\
\bottomrule
\end{tabular}
}
\end{table}
\subsection{Scaling to Larger Masked Vision Models}
To validate the applicability of the proposed ALTA to larger masked vision models, we conduct experiments by scaling the architecture to ViT-L and adopt ALTA for adaptation.
As shown in Table~\ref{table:scaling}, the results show that the ViT-L model achieves performance comparable to the default ViT-B, indicating that the proposed ALTA can be seamlessly integrated with the larger model without compromising effectiveness.
These experiments indicate that ALTA is well-suited for larger pretrained foundation models, highlighting its strong generalizability across different scales.
\begin{table}[h]
\caption{
The analysis of scaling up the masked vision model and applying the proposed ALTA to the larger vision model.
Performances of retrieval tasks are presented.
}
\label{table:scaling}
\centering
\renewcommand{\arraystretch}{1.2}
\resizebox{\linewidth}{!}{
\begin{tabular}{c|cccccc|ccc}
\toprule
\multirow{3}{*}{Method} & \multicolumn{6}{c|}{CheXpert 8$\times$200}                                        & \multicolumn{3}{c}{CheXpert 5$\times$200} \\ \cline{2-10} 
                        & \multicolumn{3}{c|}{Image$\rightarrow$Image}        & \multicolumn{3}{c|}{Text$\rightarrow$Image} & \multicolumn{3}{c}{Image$\rightarrow$Text}    \\ \cline{2-10} 
                        & P@5  & P@10 & \multicolumn{1}{c|}{P@50} & P@5       & P@10     & P@50     & P@5       & P@10      & P@100     \\ \hline
MRM (ViT-B)                     & 26.5 & 25.9 & \multicolumn{1}{c|}{23.3} & -         & -        & -        & -         & -         & -         \\
MRM (ViT-L)             & 26.8 & 26.1 & \multicolumn{1}{c|}{23.2} & -         & -        & -        & -         & -         & -         \\ \hline
Our ALTA (ViT-B)               & 51.3 & 47.1 & \multicolumn{1}{c|}{38.7} & 64.5      & 65.8     & 53.0     & 56.8      & 55.4      & 46.8      \\
Our ALTA (ViT-L)        & 50.3 & 47.9 & \multicolumn{1}{c|}{38.7} & 68.5      & 64.3     & 52.8     & 57.2      & 55.3      & 46.0      \\
\bottomrule
\end{tabular}
}
\end{table}
\subsection{Implementation Details}
\label{subsec: implementation details}
\noindent\textbf{For the proposed ALTA training.}
Our code is implemented using PyTorch 2.0.0~\cite{paszke2019pytorch}.
The training process of ALTA is carried out on 4 GeForce RTX 3080Ti GPUs for a maximum of 30 epochs.
The training is halted when the validation score does not increase for 10 consecutive epochs, taking about 10 hours.
The batch size is 96, requiring 12 GB of memory for each GPU.
AdamW~\cite{loshchilov2017decoupled} is adopted as the default optimizer, with a weight decay of 0.05, $\beta_1$ of 0.9, and $\beta_2$ of 0.95.
We employ a ``warm-up'' strategy by linearly increasing the learning rate to 5.625e-5 and then decreasing it using a cosine decay schedule.
The bottleneck ratio of the adapters in both the vision and language models is set as 0.25.
When conducting LoRA fine-tuning on the NIH ChestX-ray dataset, the LoRA ranks are set to 4, 8, and 32 for the labeling ratios of 1\%, 10\%, and 100\%, consistent with previous work~\cite{lian2024less}.
The parameters of the frozen vision and language models are initialized with weights from MRM~\cite{zhou2023advancing} and BioViL~\cite{boecking2022making}, respectively.

We adopt data augmentations for radiographs and radiology reports during ALTA training.
In practice, we resize the shorter edge to 256 and center-crop to 224.
We apply random affine transformations (rotation up to $20^{\circ}$, shear up to $15^{\circ}$, horizontal and vertical translation fractions up to 0.1, and scaling factor sampled from $\left[0.95, 1.05\right]$) and color jittering (brightness sampled from $\left[0.8, 1.2\right]$ and contrast sampled from $\left[0.8, 1.2\right]$).
For radiology reports, we randomly shuffle the sentences within a report and each report is truncated to 128 tokens. 
ViT-B is adopted as the vision encoder and a light-weight ViT encoder is used as the image decoder in line to MRM~\cite{zhou2023advancing}.
We use BERT-base as the language processor, where the tokenizer and networks are inherited from BioViL~\cite{boecking2022making}.

\noindent\textbf{For evaluation in downstream tasks.}
The fine-tuning experiments on NIH ChestX-ray and MS-CXR-T are conducted using a single GeForce RTX 3080Ti GPU, where AdamW is used as the default optimizer to minimize cross-entropy loss, with a weight decay of 0.05, $\beta_1$ of 0.9, $\beta_2$ of 0.999, and a batch size of 64.
The initial learning rate is selected from 3e-3, 5e-3, and 1e-3, according to the performance on the validation set.
A ``warm-up'' and the cosine decay schedule is also adopted here.
Other evaluation tasks do not require further tine-tuning, and inferences are conducted directly.
For retrieval tasks, vision and language representations are obtained using the image encoder and language processor after ALTA training.
For zero-shot classification tasks on RSNA Pneumonia, MS-CXR-T, and RadNLI, we perform ten-fold cross-validation while tuning the threshold with a step size of 0.005 on the validation set.
Different from ALTA training where we mask 75\% of the image patches, all image patches are kept for retrieval and classification.
The results of comparative methods are obtained either from the original papers or by conducting inferences based on the official releases.
\section{Conclusion and Discussion}
In this paper, we introduce an efficient alignment method for medical vision-language matching tasks by adapting the pretrained masked vision models within a parameter-efficient framework.
The information consistency between temporal-multiview radiographs and radiology reports is enhanced by training with records reorganized in a temporal-multiview manner, thereby further promoting vision-language alignment.
Experimental results show that ALTA outperforms previous counterparts by large margins in vision-language matching tasks and further enhances visual and language understanding.
To achieve the performance, ALTA requires fewer than 8\% of the trainable parameters, and its computational demand is less than 1/5 compared to MRM pretraining.
Furthermore, we validate the applicability of the proposed ALTA to larger masked vision models (ViT-L).
As ALTA is a parameter-efficient training method without heavy architecture-dependent designs, it is expected to generalize effectively to larger vision and language foundation models, where parameter-efficient methods yield more significant impact.
Establishing model scaling laws for medical vision-language alignment is a valuable direction, and we aim to design more efficient parameter-efficient alignment methods that work well with scalable medical vision-language models.
%
%
\bibliographystyle{IEEEtran}
\bibliography{tmi}

\end{document}